**Modular, On-Site Solutions with Lightweight Anomaly Detection for Sustainable Nutrient Management in Agriculture**

Cohen, Abigail R.[1]; Sun, Yuming[1]; Qin, Zhihao[2]; Muriki, Harsh S.[3]; Xiao, Zihao[4]; Lee, Yeonju[5]; Housley, Matthew[6]; Sharkey, Andrew F.[1]; Ferrarezi, Rhuanito S.[6]; Li, Jing[5]; Gan, Lu[7]; Chen, Yongsheng[1]*

[1] School of Civil and Environmental Engineering, Georgia Institute of Technology, North Avenue, Atlanta, GA 30332

[2] School of Electrical and Computer Engineering, Georgia Institute of Technology, North Avenue, Atlanta, GA 30332

[3] School of Interactive Computing, Georgia Institute of Technology, North Avenue, Atlanta, GA 30332

[4] School of Computer Science, Georgia Institute of Technology, North Avenue, Atlanta, GA 30332

[5] School of Industrial and Systems Engineering, Georgia Institute of Technology, North Avenue, Atlanta, GA 30332

[6] Department of Horticulture, University of Georgia, Athens, GA, United States

[7] School of Aerospace Engineering, Georgia Institute of Technology, North Avenue, Atlanta, GA 30332

* Corresponding Author, Yongsheng.chen@ce.gatech.edu

# ABSTRACT

Efficient nutrient management is critical for crop growth and sustainable resource consumption (e.g., nitrogen, energy). Current approaches require lengthy analyses, preventing real-time optimization; similarly, imaging facilitates rapid phenotyping but can be computationally intensive, preventing deployment under resource constraints. This study proposes a flexible, tiered pipeline for anomaly detection and status estimation (fresh weight, dry mass, and tissue nutrients), including a comprehensive energy analysis of approaches that span the efficiency-accuracy spectrum. Using a nutrient depletion experiment with three treatments (T1-100%, T2-50%, and T3-25% fertilizer strength) and multispectral imaging (MSI), we developed a hierarchical pipeline using an autoencoder (AE) for early warning. Further, we compared two status estimation modules of different complexity for more detailed analysis: vegetation index (VI) features with machine learning (Random Forest, RF) and raw whole-image deep learning (Vision Transformer, ViT). Results demonstrated high-efficiency anomaly detection (73% net detection of T3 samples 9 days after transplanting) at substantially lower energy than embodied energy in wasted nitrogen. The state estimation modules show trade-offs, with ViT outperforming RF on phosphorus and calcium estimation ($R^2$ 0.61 vs. 0.58, 0.48 vs. 0.35) at higher energy cost. With our modular pipeline, this work opens opportunities for edge diagnostics and practical opportunities for agricultural sustainability.

# 1 INTRODUCTION

Agriculture requires sizable amounts of energy-intensive inputs (e.g., nutrients).[1,2] The industrial fixation of nitrogen (N) for fertilizers is an energy-intensive process that consumes 35–50 MJ per kilogram.[3,4] However, the efficiency of this applied nitrogen is low, with a global average N use efficiency (NUE) of only 46%.[5] This means more than half of applied N remains in agricultural discharge waters, which contributes to eutrophication downstream and generates 65% of global emissions of nitrous oxide ($N_2O$), a potent greenhouse gas.[6,7] Improved nutrient management can be achieved through precision agriculture (PA), and automation can save time and labor expenses.[8–10] PA depends on efficient monitoring systems capable of quickly detecting problems, backed by the necessary computing infrastructure to continuously evaluate and optimize. However, if detection accuracy does not prevent meaningful crop loss, the system is too computationally intensive to offset efficiency gains, or the data infrastructure cannot be supported in resource-constrained environments, sustainable PA will not scale. With the diverse range of environmental parameters and highly variable biological responses inherent to agriculture, increasingly complex management systems could lead to higher computing and equipment costs. For these reasons, efforts should be made to create efficient, fit-for-purpose models that work in edge environments.

Non-destructive, dynamic plant monitoring has the potential to allow growers to maximize productivity, minimize resource use, and adhere to effluent standards without relying on heuristic-based or threshold-based simulations that may still result in luxury conditions.[11–13] To these ends, imaging [e.g., red, green, blue (RGB) and multispectral imaging (MSI)] and modeling [e.g., machine learning (ML)] show promise in direct, non-invasive phenotyping. Whole images can be analyzed directly for color and shape features, or reflective intensity features such as vegetation indices (VIs) can be easily extracted. RGB-VIs can monitor plant growth by relating green bands with red or blue, while MSI can improve estimation by measuring bands beyond the visual spectrum [e.g., Red Edge (RE), Near-Infrared (NIR) and Short-Wave Infrared (SWIR)]. VIs have been used to assess field crop biomass, water status, among others using UAV and satellite imagery.[14–17]

Once images are captured, modeling approaches typically involve dimensionality reduction [e.g., principal component analysis (PCA) or band filtering] followed by traditional regression or ML.[18–21] More recently, given advancements in graphical processing units (GPUs) and superior computing infrastructure, scientists have also applied deep learning (DL) to specific extracted features or whole images.[22–25] However, the resource intensity required for DL can be prohibitive for deployment in resource-constrained environments. Additionally, most nutrient analyses focus either on classification,[26–29] estimating instantaneous phenotypes,[21,30] or constructing time-series diagnostic curves for a single nutrient.[31] Moreover, wide biological variance in plant response variables (RVs) (e.g., tissue nutrient concentration, mass) can make instantaneous diagnosis challenging,[32,33] meaning growth must be evaluated against the desired evolution of parameters to support management or automation. Therefore, control optimization requires models that can make sense of biologically variable time-series data to support automated decision-making.[34,35]

While these studies advance biofeedback-enabled management, analysis bridging resource intensity and nutrient management remains limited. Computing costs and energy use increase with model complexity,[36] and the rapid pace of adoption of energy-hungry large language models (LLMs),[37] the sustainability of AI has come under scrutiny.[38,39] Not only do different modeling phases (i.e., training and inference) have variable energy demands (and by extension water use and $CO_2$ emissions), but different frameworks (e.g., TensorFlow and Pytorch) differ in impact intensity.[40] When dealing with high-dimensionality MSI data, there can also be trade-offs between performance, interpretability, complexity, and the scale of training samples, with simpler models sometimes outperforming more complex ones.[41] While some recent reviews show the possible benefits of AI in agricultural resource use,[42,43] the net impacts remain understudied with respect to nutrient management. Moreover, recent shifts towards edge computing, TinyML, and distributed microcontroller deployment require critical evaluation of model approaches and the development of efficient, modular, task-specific models for resource-constrained environments.[44,45]

This study provides a framework for efficient, adaptable, non-destructive nutrient intervention through anomaly detection, while evaluating the accuracy and efficiency of nutrient status modules. We introduce a tiered pipeline for MSI nutrient management, integrating a lightweight autoencoder (AE) for direct anomaly detection to circumvent labor-intensive spot checks followed by two potential state estimation modules ranging in complexity (random forest, RF, and vision transformer, ViT). To highlight performance and environmental impact trade-offs in these approaches, we evaluate accuracy, computation time, and energy use for each module. This integration of advanced imaging, real-time data, and established analytical methods provides a nuanced understanding of nutrient management modeling, filling the critical gap in understanding the relative impacts of model complexity compared to the potential gains of nutrient use minimization.

## 2 MATERIALS & METHODS

### 2.1 Experiment Overview

To aid in the evaluation of nutrient management modeling approaches, we conducted a pilot-scale imaging experiment, testing the limits of N removal efficiency (NRE, applied N minus effluent N), which also provided rich imaging and plant growth data.

#### 2.1.1 Plant Cultivation and Sampling

An experiment was conducted using a deep-water culture (DWC) system in a glass-covered greenhouse at the University of Georgia (College of Agriculture and Environmental Sciences, Department of Horticulture, CEA Crop Physiology and Production Laboratory) during Fall, 2024. Rex lettuce (*Lactuca sativa*) was grown for 30 days after transplanting (DAT) using a depletion fertilization strategy, where water was replenished as needed, 2-3 times per week. Three treatment groups with successively diluted nutrient solution concentrations, 100% (control, T1), 50% (T2), and 25% (T3) were prepared in triplicate (216 plants per treatment) using a modified Sonneveld solution (MSS)[46] recipe as the control (S1.1).

Plant tissue analysis was performed to provide ground-truth (GT) labels by Waters Agricultural Laboratories (Camilla, GA). Five whole-head samples (cut just above the substrate-bound root ball) from each of the nine tanks were sampled on days 11, 14, 18, 21, 23, 25, 26. Shoot samples were weighed to obtain fresh weight (FW) and packaged whole for analysis. Samples were analyzed for dry weight (DW), total carbon, N, P, potassium (K), calcium (Ca), magnesium (Mg), and sulfur (S). Water samples from the start and end of the experiment were analyzed by Waters Agricultural Laboratories (Camilla, GA). Further details of the growth conditions, tissue analysis, and water analysis can be found in supporting documentation (S 1.1).

*2.1.2   Image Acquisition & Computing System*

Plant growth was continuously monitored using a WIFI-connected (Simple Mobile Moxee 4G Mobile Hotspot, Simple Mobile; Miami, USA) imaging system (RAYN Vision Systems, Middletown, USA). Overhead multispectral images were captured every hour beginning at 10 pm (eight images per night) using five MSI cameras from 4 days after transplanting (DAT), for a total of 880 overhead images. The MSI output included 1,280 x 800-pixel images with a depth of 10 channels (blue, cyan, green, amber, red, deep red, far red, NIR-850 and NIR-940). A diagram and photos of the overhead imaging setup can be found in the supporting information (SI 1.4).

The computing system used in this experiment deployed two NVIDIA RTX 6000 Ada Generation graphical processing units (GPUs), each equipped with 48 GB of VRAM. The GPUs were paired with a 64-core AMD Ryzen Threadripper Pro 7985WX CPU, which facilitated masking, resizing, and data conversion.

The entire system ran on Linux Ubuntu 22.04.5 LTS using Python 3.9.19, leveraging CUDA 12.4 for GPU acceleration to optimize performance during model training and inference.

*2.1.3   Preprocessing & Vegetation Indices*

Following data collection, overhead images were first segmented to isolate each lettuce sample image using the Segment Anything Model (SAM).[47] As optimized, edge-enabled segmentation was out of scope for this study, an automated approach was selected for model proof-of-concept. SAM is an automated "zero-shot" tool developed to segment natural images that maintains high accuracy without the need for post-training or fine-tuning.[48] Images were converted into pseudo-RGB images, and then usable plant samples were identified (selecting only complete samples that were not cut off in the frame) and labeled for each. Next, SAM was applied to each overhead image using point-based input prompts to generate individual binary masks for each lettuce ID, which were then applied to MSI data.

Following segmentation, sample images were saved for use in ViT model and VIs were extracted along with their mean, median, standard deviation for each foliar surface for use later in the RF and AE. Three datasets were prepared for each sample and saved to the server for further analysis: (1) a whole segmented sample image for end-to-end ViT; (2) a single daily VI value for each index feature using the 11pm image; and (3) a daily average for each index-based feature. The VI-based feature set consisted of 106 VI-based input features. A table of indices and their calculations can be found in SI 1.6 along with our VI calculation pipeline details.

### 2.1.4 Tiered Nutrient Monitoring

Our study employed a multidimensional approach that included a VI-based early-warning protocol with an AE architecture trained on healthy trajectories to flag deficient trajectories, bypassing continuous state estimation in resource-constrained situations. The AE was followed by two alternative state estimation modules that ranged in complexity from ML (RF) to DL (ViT) for state estimation. In parallel, a high-resolution RF-AE was proposed for change point detection (CPD) in estimated RV trajectories. A control block diagram for the parallel pipeline can be found in Figure 1.

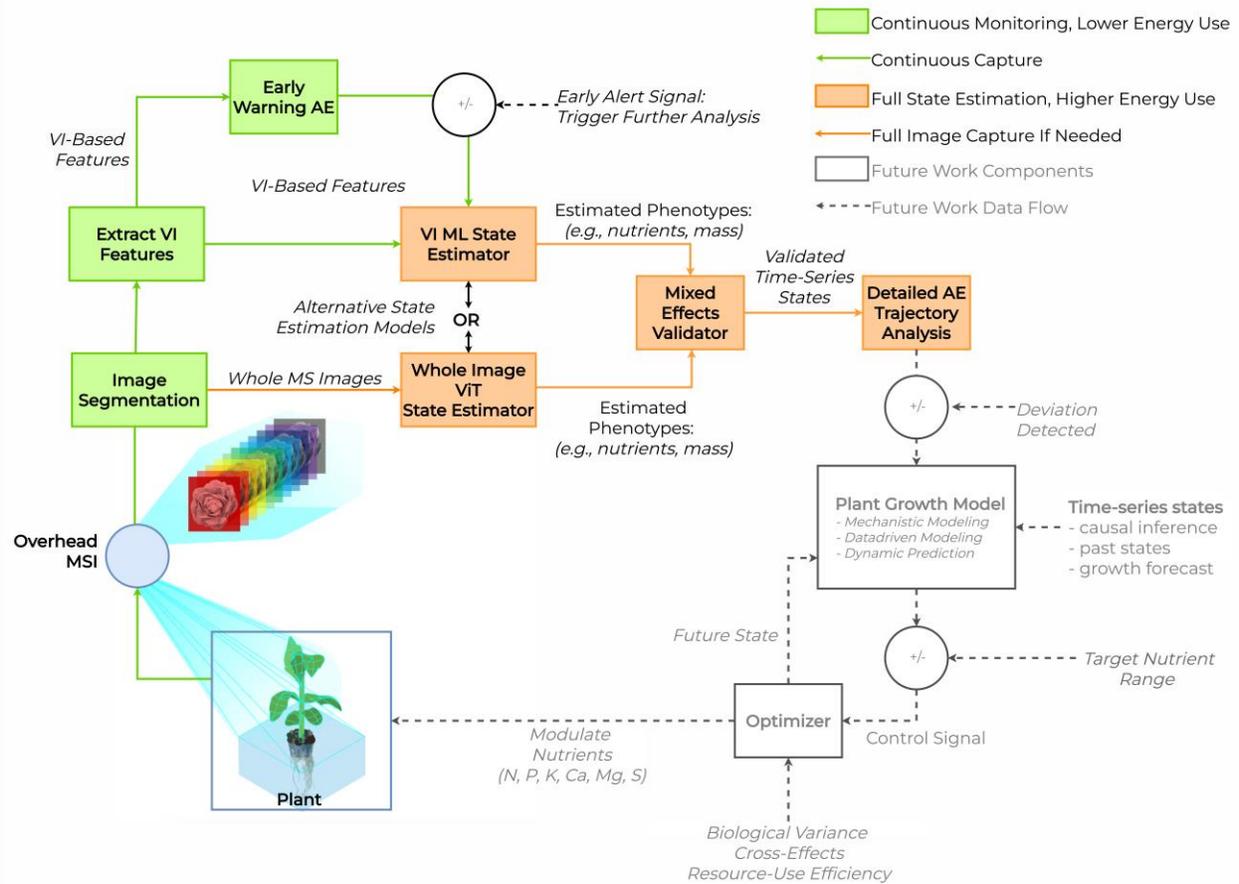

*Figure 1: Nutrient monitoring system control block diagram with three parallel pathways: (1) lightweight continuous early-warning system using autoencoder (AE) on vegetation index (VI)-based features; (2) machine learning model (random forest, RF) for state estimation using VI-based features; and (3) whole image deep learning vision transformer (ViT) state estimation. To demonstrate automation opportunities, state estimation results pass through AE for detailed trajectory analysis.*

This tiered approach, including the AE anomaly detection and state estimation module energy analysis, along with a system-level integration and validation approach using a mixed effects model (MEM), provides a framework for enhanced nutrient management sustainability in a range of operating environments.

## 2.2 Anomaly Detection using Autoencoder (AE)

One of the core contributions of this study is the demonstration of a modular AE anomaly detection approach. AE is a neural network designed to learn latent representations of input data through unsupervised learning to optimize the reconstruction of the input.[49] The AE architecture contains two primary components: (1) an encoder learns to compress input data into low-dimensional latent representations, and (2) a decoder reconstructs the original data from these learned representations. Through this encoder-decoder framework, AEs capture underlying data patterns and structures of trajectories they have seen before and struggle to reconstruct new, anomalous trajectories.

The AE proposed in this study adapts a simple autoencoder architecture, which consisted of a fixed-dimension 64-32-64 encoder-decoder, using the nn.module in PyTorch.[50,51] To illustrate its adaptability, two implementations of the AE were evaluated: (1) an early-warning module using VI-based features extracted from one MSI image of each sample per day, starting at 4 DAT. All 106 VI-based features were analyzed using the AE, and the top 5 were selected for visualization. (2) A high-resolution module using RF-estimated response variable trajectories as inputs to the AE model, generated using the daily-averaged VI-based feature dataset (not used for RF training) to produce pseudo-labels.

The models were each trained over 100 epochs on exclusively "healthy" T1 trajectories, VI-based feature trajectories for the early warning module or RF-estimated trajectories for the high-resolution module, using the Adam optimizer with a learning rate of 0.001. During inference, growth trajectories of lettuce heads ($y_t$) at time t were reconstructed as $\hat{y}_t$ with a reconstruction error threshold of 1.5 times the mean reconstruction error observed in the training dataset at the final training epoch, serving as the indicator for anomaly detection.

Windows from 6-22 days long, starting at 4 DAT, were used to evaluate the optimal length of time needed for the model to differentiate between healthy (T1) and deficient (T2) or highly deficient (T3) trajectories. The models were each trained on the first 40 (or as many as available given destructive sampling) of the T1 trajectory segments of the same length, and then each window was tested on the remaining T1 trajectories along with the T2 and then T3 trajectories, respectively, to evaluate the net detection accuracy (true detection rate minus false detection rate). Performance metrics evaluated for the AE included the net detection rate (true detection minus false detection) and GPU energy consumption.

## 2.3 Deeper Analysis: State Estimation Module Options

Nutrient management intervention and automation requires a balance between accurate state estimation and efficient deployment. For the state estimation modules used in this study, RF and ViT models were evaluated using the single 11 pm image (for ViT) and VI calculation of this image (for RF). The dataset was divided for both models with the same training, validation, and testing data splits. For each approach, images were divided into two categories: (1) labelled images with ground truth (GT) analysis; and (2) unlabeled images.

To evaluate performance while maintaining tank and treatment representation and avoiding data leakage, a manual 27-fold cross-validation scheme was developed and replicated for both RF and ViT. Each fold represented a unique permutation of tanks, maintaining two tanks from each treatment class for testing and the remaining for training (80%) and validation (20%) to ensure consistent and comprehensive coverage of the dataset. The test set contained only labeled data. Unlabeled images from each training and validation tank were assigned pseudo-labels using the average RV values from their respective tanks using the GT values of the 5 destructive samples from that DAT.

Finally, Weights & Biases[52] was integrated into the pipeline to enable real-time tracking of training metrics and visualizations of model performance over time. This tool also facilitated reproducibility by maintaining a detailed record of each experiment's configuration and results.

### 2.3.1 Vegetation Indices with Machine Learning

Among traditional ML approaches, random forest (RF) produces reliable predictions using non-linear decision trees, and when used for regression analysis, RF is robust to multicollinearity.[53,54] Compared to other ML approaches, RF has been successful at a variety of phenotyping tasks and has been found to be fast with relatively low computational intensity.[55,56] RF was used for two tasks in this pipeline. First, it served as the feature selection module. Second, it was used to determine instantaneous phenotypes for each lettuce sample RV (FW, DM, and tissue N, P, K, Ca, S, and Mg) with VI-based input features. All modeling, including the feature selection procedures, was implemented using the scikit-learn library in Python. The modeling workflow was applied independently to each RV target.

RF feature selection was conducted in three stages within each fold: (1) recursive feature elimination (RFE); (2) removal of low-importance features; and (3) correlation pruning. Additional details of the feature selection can be found in SI 2.1.1. Selected features were aggregated across the 27 folds and ranked based on frequency and median importance, with the top 20 features selected as the fixed feature set. Each fold was subsequently rerun using these 20 features with the tuned hyperparameters identified previously. In this final training, the selected features underwent a final round of correlation pruning with the same criteria.

The final model was retrained using the pruned feature set and corresponding hyperparameters, and predictions were generated for the test set. SHAP (SHapley Additive exPlanations) was used to interpret the final RF models and quantify the contribution of each selected vegetation index to the predictions. In addition, out-of-fold (OOF) test predictions were collected by aggregating the predicted values from all test folds, enabling an overall out-of-fold evaluation of model performance.

### 2.3.2  Whole Image Analysis with Deep Learning

The second state estimation model selected for analysis was a DL vision transformer (ViT) using multi-task learning to predict all RVs simultaneously. ViT was selected for its reputation as both a state-of-the art computer vision approach and a highly complex model that has not been widely explored for use in nutrient state estimation[57] but shows promise for precision agriculture.[58] It was also selected over more commonly deployed architectures such as convolutional neural networks (CNN) like ResNet[59] due to ViT's ability to make use of spatial and spectral relationships using self-attention mechanisms. ViTs processes images as sequences of patches, unlike CNNs, which focus on local spatial features, which makes ViTs adept at recognizing global relationships across whole images.

Using the labelled and unlabeled data in the same manner as the RF, the ViT model was adapted to use raw multispectral images with a suitable feature space, using all 10 spectral channels afforded by the MS imager from the original 3-channel input. The ViT had 2 main components: (1) a ViT component (implemented using vit_pytorch),[60] which processed patches through multiple self-attention layers and discovers spatial and spectral patterns to estimate instantaneous tissue nutrient concentration and mass[61]; and (2) a fully connected layer that converted the output of the ViT followed by rectified linear unit (ReLU) activation functions which reduce the dimensionality and introduce non-linearity into the dataset for more accurate predictions. More details of the model architecture can be found in SI 2.1.2.

## 2.4 Performance Metrics and Validation

Efficient, biologically accurate, modular models are becoming increasingly important for edge computing and deployment in resource-constrained environments. To evaluate the different modules in the tiered pipeline in this context, we measured accuracy and GPU energy usage for each module. We then evaluated the biological validity of results based on treatment effects using mixed effects modeling.

### 2.4.1 State Estimation Accuracy Metrics

To evaluate and compare the two state estimation modules, we calculated the coefficient of determination ($R^2$) and root mean squared error (RMSE) for each fold, training and inference time, and energy usage for model training/validation and inference. Per-nutrient $R^2$ and global $R^2$ values were calculated for each fold, from which the average and standard deviation across the manual cross validation 27 folds were determined. RMSE was calculated to measure the prediction accuracy of each response variable estimation at each time point, indicating the average magnitude of the model's prediction errors.

### 2.4.2 Energy Measurement Methodology

The computational efficiency of each module was calculated using GPU power monitoring for benchmarking purposes, running each model on one of the two GPUs (GPU 0 or 1) to streamline monitoring and documentation. For longer-running models (ViT), Weights & Biases was used to extract power consumption data (watts) in real-time directly from the NDVIA-smi, which uses

NVIDIA Management Library (NVML), an interface for monitoring and managing NVIDIA GPU states.[62] For the very short runs of the RF and AE models, NVML was accessed directly to monitor average power consumption over the course of the runs, which returns the power draw (±5 W) averaged over one second. To determine uncertainty for power use, the larger value was used (standard deviation or 5 W). Note: since the RF was developed using the Ski `scikit-learn` library, which uses CPU exclusively, we needed to rerun the RF using the `cuml.RandomForestRegressor` library to force it to run on the GPU for even hardware comparison.

Power consumption was multiplied by the total runtime to calculate energy consumption for the run (Wh).[52,63] GPU energy consumption and computing time were recorded during both the training and inference phases, where each training phase included all steps involved in model training, validation, and hyperparameter tuning (excluding the RFE for RF), while the inference covered final prediction on the entire available dataset (with exception of the AE, which only had training and testing sets). For per-sample estimates, training and inference energy was calculated for each sample and multiplied by run duration to find an average per-sample consumption. Additional energy considerations such as RAM and CPU use were out of scope in this study.

### 2.4.3 Mixed Effects Modeling to Evaluate Treatment Effects

Following estimation, mixed effects modeling (MEM) was deployed to evaluate the source of variance in growth response and to validate the underlying assumptions. MEM was applied to both GT and RF-estimated trajectories using a linear MEM model implemented by python (mixed.lm).[64,65] The model used a basic two-tier nested structure with three compartments: treatment (fixed effect), tank-within-treatment, and residual effects (random effects, respectively).

The MEM served three major functions in our analysis: First, for experimental design validation, confirming treatment dominates variance in GT data. Next, the MEM served as a state estimation fidelity assessment to validate the biological signal reliability of our estimation models was preserved in our estimated RV trajectories. Finally, we used the analysis as an intervention trigger validation to confirm that the anomaly detection signal was related to treatment.

## 3  RESULTS & DISCUSSION

A shift from centralized computation towards soft sensors and edge computing in agriculture will require a thorough evaluation of model complexity, efficacy, and efficiency. Our analysis provides a novel evaluation of nutrient management model energy use in the context of avoided embodied energy from nutrient waste, proposing a tiered approach to nutrient management that includes efficient anomaly detection followed by deeper trajectory analysis.

The results from this experiment are presented in three parts: (1) individual module performance and energy use, including the early warning system using AE model on the raw VI-based features and a performance comparison of the two state estimation modules (RF and ViT); (2) an integrated system using AE on the RF-estimated trajectories to demonstrate that modular, fit-for-purpose DL trajectory analysis can be deployed at multiple stages; and (3) analysis and discussion of how energy use might scale for industrial implementation as compared to the potential to offset embodied nitrogen use from abated waste.

### 3.1  Individual Module Performance

#### 3.1.1  Lightweight Anomaly Detection using VI-AE

Early warning of nutrient application issues can save operators significant time and labor expenses. First, our results using the VI-AE demonstrate that selecting the appropriate VI is crucial to the model's success, and that even the top features evolve in their detection efficacy over the course of the experiment (Figure 2).

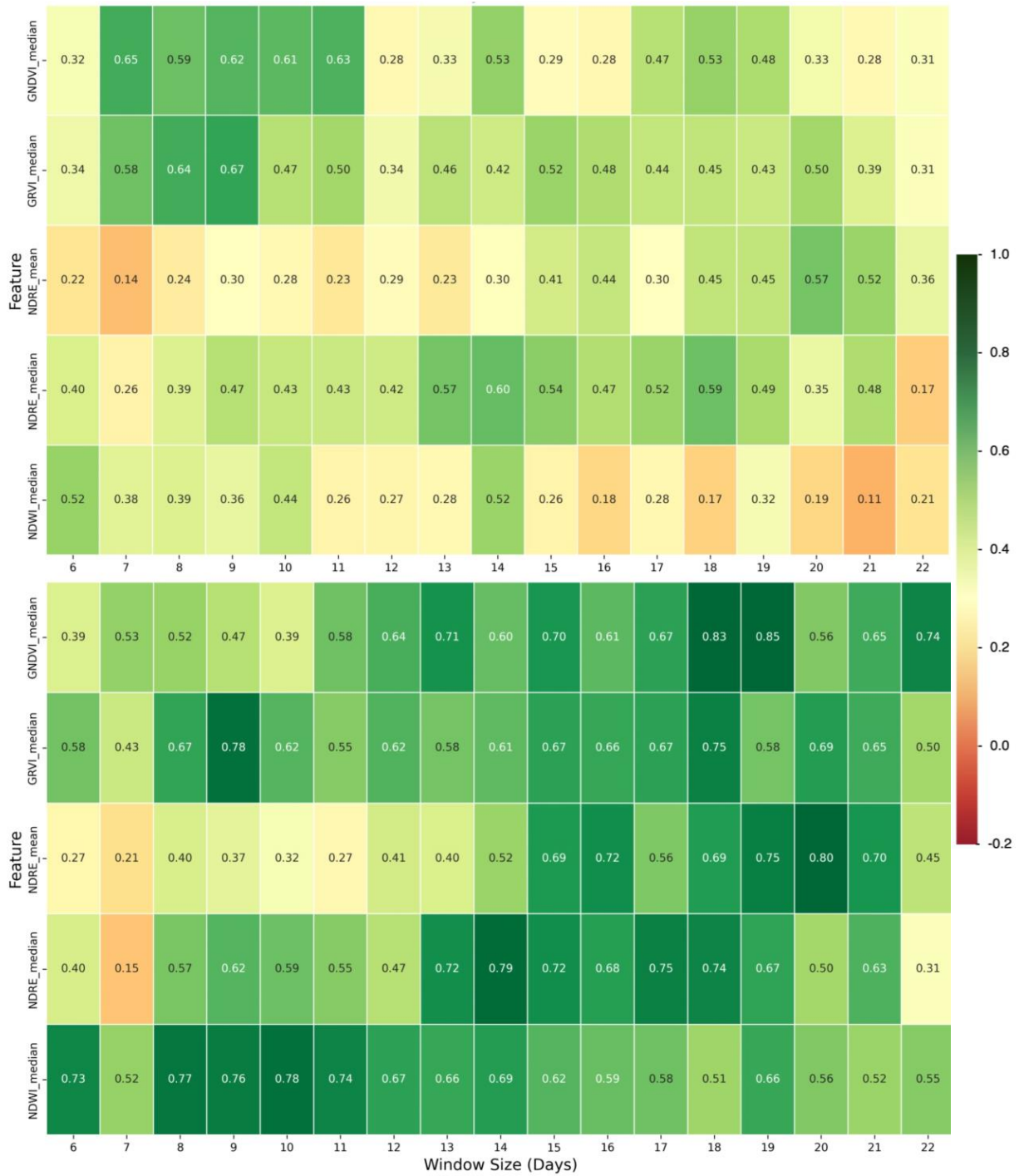

*Figure 2: Net detection rate (true positive minus false positive) heat map of top 5 features for the early warning autoencoder for T2 (top) and T3 (bottom) by day from 4-11 DAT through 4-25 DAT.*

Different VIs may be better suited for monitoring anomalies at different growth stages, with NDWI median showing most promise for early growth periods (with 71% average net detection of T3 until window length 12) and GNDVI median for later growth (73% average net detection of T3 for the longest five windows). Although net detection rates of T2 were lower than T3, with a 65% net detection using GNDVI for a 7-day window and a 67% net detection using GRVI with a 9-day window, significant losses could be averted using this early warning protocol prior to deeper analysis.

Early plant growth can be especially sensitive to nutrient and water stress, when small leaves are thin and will allow shorter wavelengths to be reflected by biochemical constituents, making indicators like NDWI promising for the detection of differences.[66] Similarly, GRVI is sensitive to early changes in green color,[67] which may explain why it shows promise as an early detection indicator, reaching 78% net detection of T3 and 67% net detection of T2 by 9 DAT. By contrast, studies exploring the use of GNDVI have found it to be an especially successful indicator of plant health during intermediate and later growth stages when the canopy is dense, when other metrics, such as the NDVI may saturate.[68] This may explain its increasing detection accuracy using the AE early warning system. Finally, the red edge channel shows promise in anomaly detection [31]: its use in the calculation of NDRE may explain the success of NDRE as a feature for differentiation between deficient and healthy trajectories as plants reach maturity, remaining above 50% net detection of T3 from 19-25 DAT, peaking at 80% at 24 DAT.

### 3.1.2 Performance of State Estimation Module Options

Nutrient management intervention and automation require a balance between accurate state estimation and efficient deployment. In this section, we present the accuracy and GPU energy use of two potential state estimation models, ranging in complexity from ML (RF) to DL (ViT). Table 1 displays the results for the state estimation models, including average $R^2$ and RMSE across the 27 folds along with time and energy consumption for training and inference.

Table 1: State Estimation Model Performance. Best $R^2$ scores are bolded for each response variable.

| Metric | Overall | FW (g) | DM (g) | N (%) | P (%) | K (%) | Mg (%) | Ca (%) | S (%) |
|---|---|---|---|---|---|---|---|---|---|
| | | | | Random Forest on VI-Based Features | | | | | |
| Avg. $R^2$ ± SD | **0.69** ± 0.06 | **0.79** ± 0.06 | **0.71** ± 0.1 | **0.82** ± 0.03 | 0.58 ± 0.06 | **0.76** ± 0.04 | **0.28** ± 0.07 | 0.35 ± 0.09 | **0.55** ± 0.09 |
| RMSE ± SD | - | 24.95 ± 3.59 | 0.72 ± 0.14 | 0.49 ± 0.04 | 0.10 ± 0.01 | 0.86 ± 0.07 | 0.04 ± 0.005 | 0.17 ± 0.02 | 0.02 ± 0.003 |
| | | | | ViT on Whole MS Images | | | | | |
| Avg. $R^2$ ± SD | 0.53 ± 0.12 | 0.7 ± 0.16 | 0.29 ± 0.11 | 0.79 ± 0.15 | **0.61** ± 0.15 | 0.65 ± 0.14 | 0.24 ± 0.08 | **0.49** ± 0.15 | 0.44 ± 0.12 |
| RMSE ± SD | - | 13.3 ± 1.50 | 0.91 ± 0.06 | 0.33 ± 0.06 | 0.07 ± 0.01 | 0.87 ± 0.08 | 0.03 ± 0.003 | 0.12 ± 0.08 | 0.03 ± 0.002 |

RF outperforms ViT in $R^2$ by 6 of 8 (80%) of the phenotypes response variables with much more stable results (average $R^2$ standard deviation of 0.07 vs ViT's 0.13). Overall, the models perform similarly with respect to N, with $R^2$ of 0.79 ± 0.15 and 0.82 ± 0.03 for ViT and RF, respectively, and for P, with 0.61 ± 0.15 for ViT and 0.58 ± 0.06 for RF, and both models struggle with Mg, a commonly troublesome nutrient for image-based estimation. However, ViT significantly outperforms RF in the estimation of tissue Ca, with a 40% increase in $R^2$ (0.49 vs RF's 0.35). For context, one study deploying HSI reported $R^2$ as high as 0.94 for Ca and 0.987 for nitrate.[69] Our approach provided the opportunity to evaluate the efficacy and efficiency of more affordable equipment (MSI vs. HSI) in state estimation under field-like conditions that likely led to weaker signals for individual tissue nutrient concentrations but more closely approximate industrial deployment. We also varied the nutrient solution strength, providing a complement to existing studies that vary nutrients independently, while imaging with MSI throughout the growth period to overhead characterize whole lettuce *in vivo* at the pilot scale.

Our results also show that the standard deviation of many of the indices works more effectively than the mean or median value of the index over the foliar surface. This suggests that the distribution of the index values provides more information about the growth of the plant than the index value. This provides critical insight into spatial characteristics of nutrient deficiency: variation across the foliar surface, lost when averaging over the sample, is highly correlated with lower tissue concentrations. While ViT is typically leveraged for its ability to merge spectral and spatial information, our results demonstrate that through careful feature engineering that captures spatial characteristics combined with simple model architectures can be more effective and much more computationally efficient than complex end-to-end approaches on whole images. Additional details about feature importance can be found in the supporting information (S 3.2.1).

### 3.1.3 GPU Energy Use Across Modules

Trajectory analysis is of critical importance for MPC and PA, facilitating dynamic, tailored control of nutrient use and the optimization of yield while maintaining the possibility of integration into edge environments. There must be continuous or intermittent monitoring of growth and phenotype trajectories to indicate success or to flag that a deviation from the desirable growth trajectory necessitates input modulation. Table 2 shows the energy analysis for the AE.

*Table 2: Energy use\* for training and inference for the autoencoder model for one feature.*

|  | Average ($\mu_i$) | Standard Deviation ($\sigma_i$)\*\* | Coef. of Variation (%) |
|---|---|---|---|
| Training Power, P (W) | 65.69 | 0.30 | 0.46 |
| Inference Power, P (W) | 69.94 | 0.24 | 0.34 |
| Training Time per Sample, t (h) | 2.0 e-6 | 2.35 e-7 | 11.82 |
| Inference Time per Sample, t (h) | 2.55 e-8 | 1.74 e-9 | 6.83 |
| **Training Energy per Sample, E (Wh)** | **1.31 e-4** | **1.55 e-5** | **11.83** |
| **Inference Energy per Sample, E (Wh)** | **1.78 e-6** | **1.22 e-7** | **6.84** |

*\*Assumes comparable training and inference energy use per sample across features, of $1.31 \times 10^{-4} \pm 1.22 \times 10^{-7}$ Wh and $1.78 \times 10^{-6} \pm 1.22 \times 10^{-7}$, respectively.*
*\*\*$\sigma_E = \sqrt{[(\mu_t \times \sigma_p)^2 + (\mu_p \times \sigma_t)^2 + (\sigma_p \times \sigma_t)^2]}$*

Sample trajectories were also divided by window length to evaluate if the length of the trajectory impacted energy use, but a significant increase in coefficients of variation in per-day energy use demonstrated that window length was not a dominant driver of energy use variation.

*Table 3: Training and inference time and energy use for the state estimation module alternatives.*

| State Estimation Model | Training Time (h) | Inference Time (h) | Training Energy (Wh) | Inference Energy (Wh) |
|---|---|---|---|---|
| RF on VIs | 9.20 e-3\* | 6.00 e-4\*\* | 1.21 ± 0.14\* | 0.05 ± 3.90 e-3\*\* |
| ViT on Whole MSI Images | 0.673 ± 1.87 e-4 | 0.167 ± 2.5 e-4** | 66.53 ± 3.37 | 11.5 ± 0.83** |

*RF energy estimates are conservative owing to RF's CPU-intensive computation; may reflect the baseline GPU energy use*
*\*RF training time and power use was recorded for one of the 27 folds (fold 18), using the entire dataset (60 test, 781 training, and 202 validation samples), run 10 times to obtain average and standard deviation power use.*
*\*\*Inference was run over the entire dataset for both models, and std uses NVML power average ± 5 W, as it is larger than the std of power use over the entire ViT run.*

Our analysis shows that increased computational intensity does not always lead to improved accuracy across all nutrients. The RF-pipeline uses $1.6 \times 10^{-5} \pm 1.1 \times 10^{-6}$ GPU energy for inference per sample with an average model $R^2$ of $0.69 \pm 0.06$. By contrast, the deep learning modeling approach modeled here (MSI-ViT) uses $3.49 \times 10^{-3} \pm 2.5 \times 10^{-4}$ Wh for inference per sample, with overall $R^2$ of $0.53 \pm 0.12$. Per sample energy comparisons across modules can be found in Table 4.

*Table 4: Performance and energy use across model complexity, including AE on raw VI-based features, RF on daily VI-based features and ViT on whole MSI images.*

| Model | Input Data | Average $R^2$ | Training GPU Energy (Wh/sample) | Inference GPU Energy (Wh/Sample) | Relative Inference Energy Intensity |
|---|---|---|---|---|---|
| **AE on VI-Based Feature** | Selected VI-based Features | N/A (anomaly detection) | 1.54 e-4\*\*\* | 1.45 e-6\*\*\* | 1 (baseline) |
| **RF on VI-Based Features** | 20 Selected VIs | 0.69 | 1.24 e-3\* | 1.60 e-5\*\* | 8.95x |
| **ViT on Whole MSI** | Whole MSI (10 channels) | 0.53 | 6.38 e-2 | 3.49 e-3\*\* | 1956x |

*\* RF training energy was recorded for one of the 27 folds (fold 18), using the entire dataset (60 test, 781 training, and 202 validation samples), and divided by this total to estimate per-sample values.*
*\*\* Inference was run over all daily images for the 4-week experiment and divided by this total to estimate per-sample values.*
*\*\*\* AE energy use calculated for 1 feature (NDWI) using average power multiplied by each time per training or testing sample trajectory and averaged.*

The results show trade-offs in efficiency and accuracy for some nutrients. For example, the MSI-ViT has 42% $R^2$ improvement for Ca over the RF (0.49 vs 0.35, respectively) but at nearly 2000 times the inference energy per sample. Given the importance of Ca dosing accuracy and its impact on tipburn, which can result in significant economic losses for growers, Ca could be managed much more effectively with this increased accuracy for Ca estimation. This would allow growers to avoid other measures, such as reducing daily light integrals to avoid this costly occurrence.[70] While the training per sample for RF was only 80% of the AE training, the inference energy used for AE is less than half that of RF.

Though the accuracy of the deep learning state estimation approaches presented here shows room for improvement in future work with hyperparameter tuning and modifications to architecture, this framework for evaluation demonstrates the need to consider the computational intensity of the modeling approach and material efficiency trade-offs within the broader context of resource intensity. Our results suggest that while complex deep autoencoder-LSTM (long, short-term memory) classification approaches can achieve high classification accuracy (94%),[29] simpler autoencoder architectures may be sufficient for early warning implementations with careful feature engineering. Future analysis should include sensitivity analysis investigating the trade-offs in model performance and energy use across different modeling approaches with modifications to architecture, training sample sizes, and hyperparameter values.

### 3.2  High-Resolution Trajectory Analysis and Validation

While status estimation is ultimately necessary for nutrient optimization, trajectory analysis is critical for fully automated modulation. Given the low inference energy required by the RF, we also deployed the autoencoder on the RF-estimated trajectories, demonstrating the applicability in automation and potential for further resolution while still using low GPU energy.

*3.2.1  Trajectory Analysis of RF-Estimated Response Variables*

To facilitate automated decision making, the autoencoder architecture shows promise for detecting differences in estimated tissue nutrient concentration trajectories, especially in more highly differentiated treatments (T3 compared to the T1 control). In the moderate-energy detection pipeline, which begins with VI feature extraction, followed by RF state estimation, the estimated response trajectories are then run through an autoencoder (AE) at increasing window lengths (6-22 days long) from 4 DAT (Figure 3).

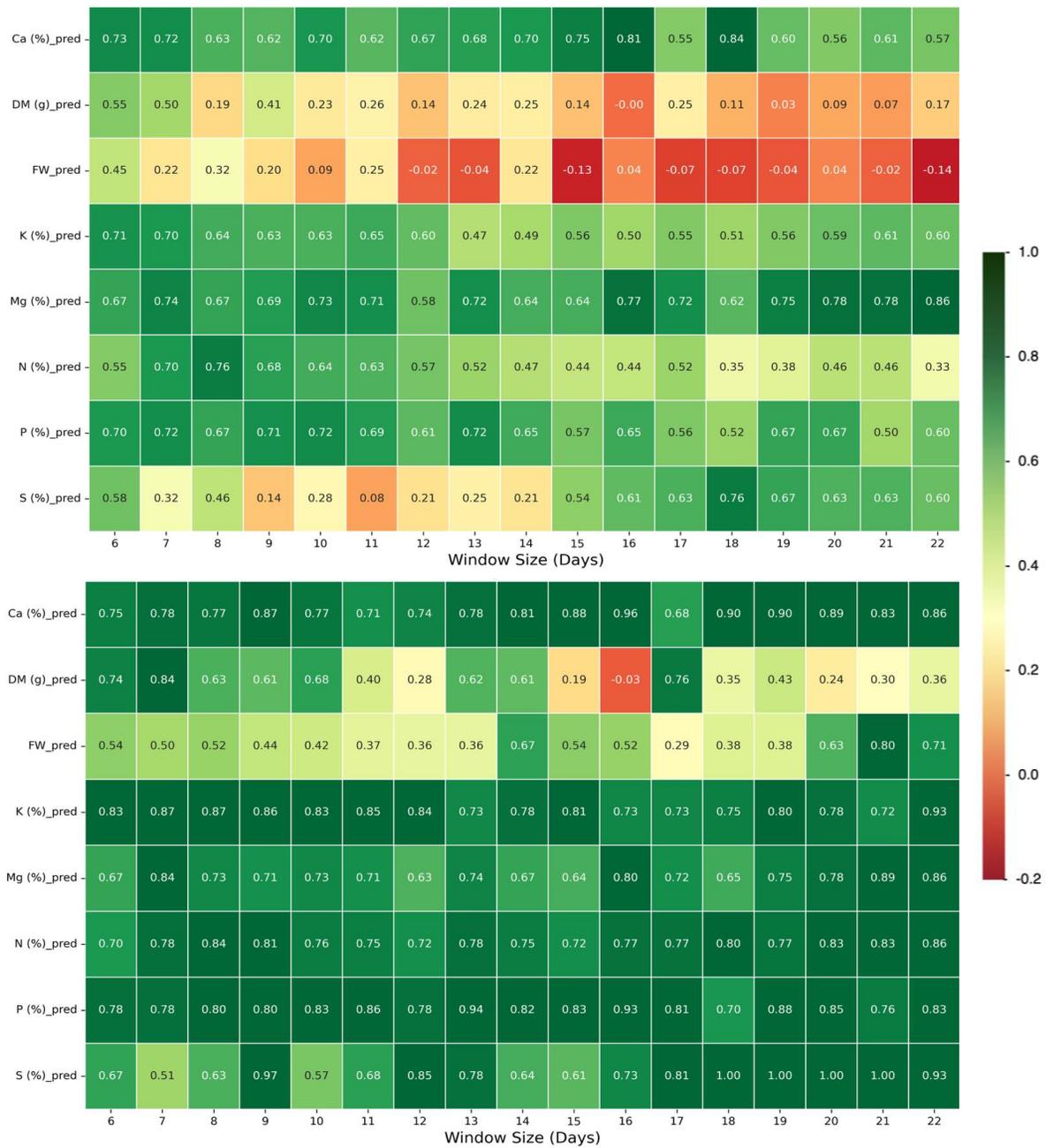

*Figure 3: Net detection rate (true positive minus false positive) for the estimated 50% (T2) trajectories (top) and estimated 25% (T3) trajectories (bottom) by response variable success rate for window sizes (6-22), with all windows beginning at 4 DAT.*

As may be expected, the RF-AE has a higher net detection rate than VIs alone, and greater success differentiating between healthy (T1) trajectories and extreme nutrient restricted treatments (T3) (Figure 3 bottom) as compared to T2 trajectories (Figure 3 top). Nutrient response variables elicit particularly high net detection of T3 treatments throughout the measured windows, from 4-10 DAT (window size 6) through 4-26 DAT (window size 22). This is especially true for Ca (which averaged 81.6%), K, N, and P. The high false detection of the T1 trajectories when tested against T2 trajectories demonstrates that the model performs better when the trajectories are highly differentiated. The weakening evolution of the T2 detection accuracy may be explained by two compounding challenges: (1) the number of training samples decreases as the window lengths increase from 6 (with 40 samples) to 22 (16 samples) with destructive sampling; and (2) the signal-to-noise ratio increases with increasing variance as plants mature.

This can be seen from the differences between S and FW detection in the later growth stages: there is strong reconstruction and net detection rate at later growth stages for T3 S trajectories (with 100% net detection by 22 DAT), even though the absolute separation between T1 and T3 trajectories is small (with mean separation of 13%), while the FW detection rate is much lower, with higher absolute separation (57% mean separation) (see estimated trajectory visualizations in SI 3.2.3). This emphasizes AE sensitivity to trajectory shape more than the values of the response variables, as it struggles with smoother, parallel curves of the weight response variables. While T3 and T2 trajectories differ in value, each show an increase that levels off in parallel over time following logistic growth.

### 3.2.2 Variance Validation using Mixed Effects Model

Following integrated state estimation and trajectory analysis, MEM was conducted to validate assumptions and assess the onset of treatment effects during the growth period. The results show that treatment dominates the variance across FW, DM, and tissue N concentration for all samples on all 7 sampling days. Apart from magnesium, treatment is the dominant source of variance for the remaining response variables on 6 of the 7 sampling days.

The treatment effects across nutrients and time (p≤ 0.05) (S3.3) reveal that treatment was a significant driver of variance for all variables on all modeled days apart from day 26 for Mg. These results also demonstrate that the biological signal of variance due to treatment is not lost in the RF state estimation model. Further, this result in estimated states' treatment effects validates the RF-AE, for which all days and windows display statistically significant treatment effects (p≤0.05) apart from Mg on 24 DAT.

## 3.3 Balancing Considerations for Resource Use in Digital Agriculture

Despite increased scrutiny facing AI's energy use, agricultural research lacks grounded benchmarking of computational approaches to evaluate against potential sustainability gains. Digital agriculture offers great promise for sustainability and material use efficiency,[71] but its implementation requires a balance between the material or energy use offsets with the computational and resource intensity of the modeling approaches used to achieve them. In this section we provide scaled GPU energy comparisons between the different modules and contextualize the relative energy use as compared to the potential mitigated embodied energy from N use reduction.

### 3.3.1 Scaling Potential Impacts of Nutrient Use Reduction

The results of our depletion experiment add to the body of evidence that plants can achieve significantly improved NRE when fed less than the industry standard.[72] While the destructive sampling design of this experiment prevented precise estimation of nutrient use efficiency (i.e., bioaccumulated N divided by applied N, or NUE), treatment-level comparisons offer promising insight into NRE and potential to reduce wasted N. The final fresh weight of the T2 plants, despite being exposed to initial N concentrations 50% of the control (T1) plants, averaged 78.9% (± 9.6%) of the control FW, with tissue N concentrations averaging 4.9% (± 0.2%) of DM, within the published thresholds for healthy tissue nitrogen levels (4.0-5.6%).[73] Moreover, the T2 plants achieved this without any remaining effluent N, showing 100% NRE, while the T1 retained 40 mg/L N in the effluent. This effluent disparity, coupled with the health and vigor achieved by the T2 plants, reveals the potential for avoided downstream impacts in addition to the upstream embedded emissions avoidance.

To contextualize the energy trade-offs at scale, we calculated inference GPU energy use and embodied energy in wasted nutrients for a facility growing 10,000 heads of lettuce over a 28-day growth period and compared them (Figure 4).

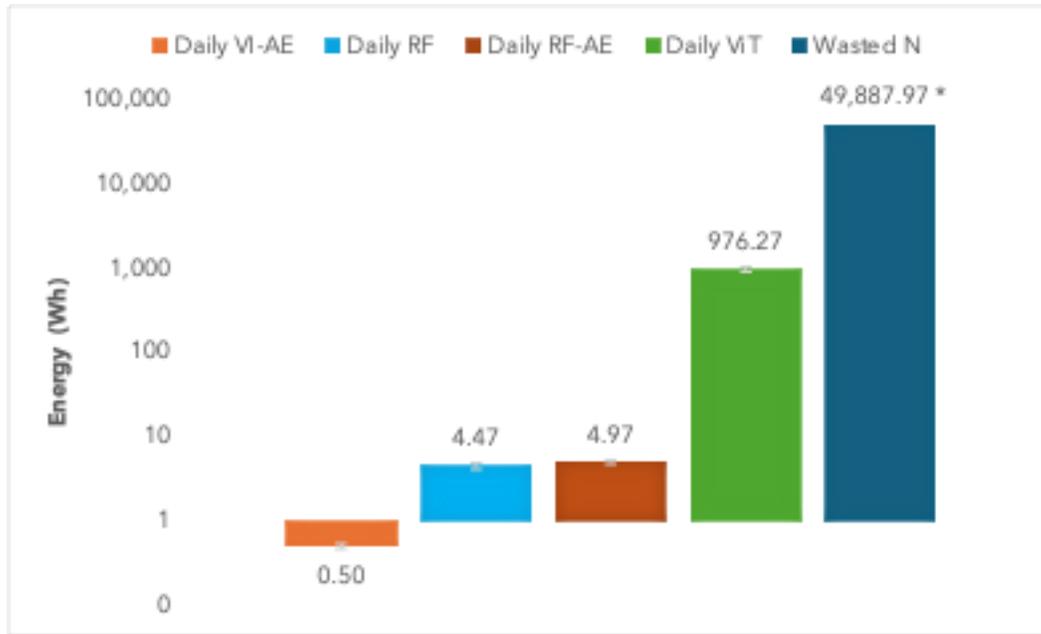

*Figure 4: GPU energy use (Wh) for 28 days of daily inference on 10,000 lettuce samples for VI-AE, RF, RF-AE, and ViT compared to embodied energy in wasted N over the same period. *Average value (11.11 kWh kg$^{-1}$ N) used from literature.[3,4] Log scale y-axis to display all values for comparison.*

It should be noted that Figure 4 uses a log scale for the y-axis to display all values in one graphic. If run every day on successively longer windows, the early-warning monitoring system (AE-VI) would require a total of 0.5 ± 0.03 Wh for the month. Since the inference GPU energy required by RF is still low (4.47 ± 0.32 Wh), a combined RF-AE system would only use 4.97 Wh over that period if run each day for a higher-resolution analysis. Meanwhile, the higher-complexity model requires significantly more inference energy under the same circumstances (MSI-ViT: 0.98 ± 0.07 kWh). **However, this computational energy consumption pales in comparison to the wasted embodied energy from excess N application.** Seeing that global NUE averages 46%,[5] our experimental scenario (with total average T1 tissue N 0.383 g ± 0.052), scaled to the same 10,000 heads of lettuce, would require 8.32 ± 1.13 kg N, of which 4.49 kg ± 0.61 N would be wasted as effluent. At 9.7-13.9 kWh per kg N, the wasted embodied energy is between 44 and 93 times the projected GPU energy used running ViT inference on MSI images over the same period. **The ViT inference GPU energy would be completely offset by a reduction of just 2% in wasted N.**

N-use optimization could yield tremendous savings in avoided embodied carbon, eutrophication burden for aquatic ecosystems, and $N_2O$ emissions from unused N in effluent solutions without compromising yield.[74] A dynamic, biofeedback-enabled nutrient monitoring system, such as the early-warning VI-AE, coupled with higher-resolution state estimation models, such as RF or ViT, could significantly reduce the harmful impacts of N mismanagement.

*3.3.2 Integrating Results for Practical Guidance & Future Directions*

Nutrient management using digital agriculture will become increasingly important for operators of both CEA and row crops. As it is not physically or economically feasible to collect samples for analysis (and farmers need all their yield for sale), anomaly detection and state estimation approaches will facilitate non-destructive, dynamic nutrient management in real-time. Here, we synthesize the tiered approach to inform the application of the proposed modular architecture (Figure 5).

First, for lower computing cost and lower-energy nutrient monitoring, a lightweight, VI-AE module demonstrates 78% net early detection of T3 by 10 DAT, which could result in significant reduction of expensive losses due to substrate supply challenges (e.g., tube clogs, uneven nutrient distribution in the grow media). Further, the approach used much less data, using a small, stored data array (VI-based feature array, 5.5 MB total for our experiment) rather than raw images (10.7 MB per image). Future implementations leveraging multiple VIs into a multivariate detection platform would likely lead to improved, robust detection over the duration of the grow out. For example, joining VI features that are more sensitive to early deficiencies such as NDWI with those that are sensitive to later changes like GNDVI along with mid-stage detection indices (NDRE) could lead to sustained detection.

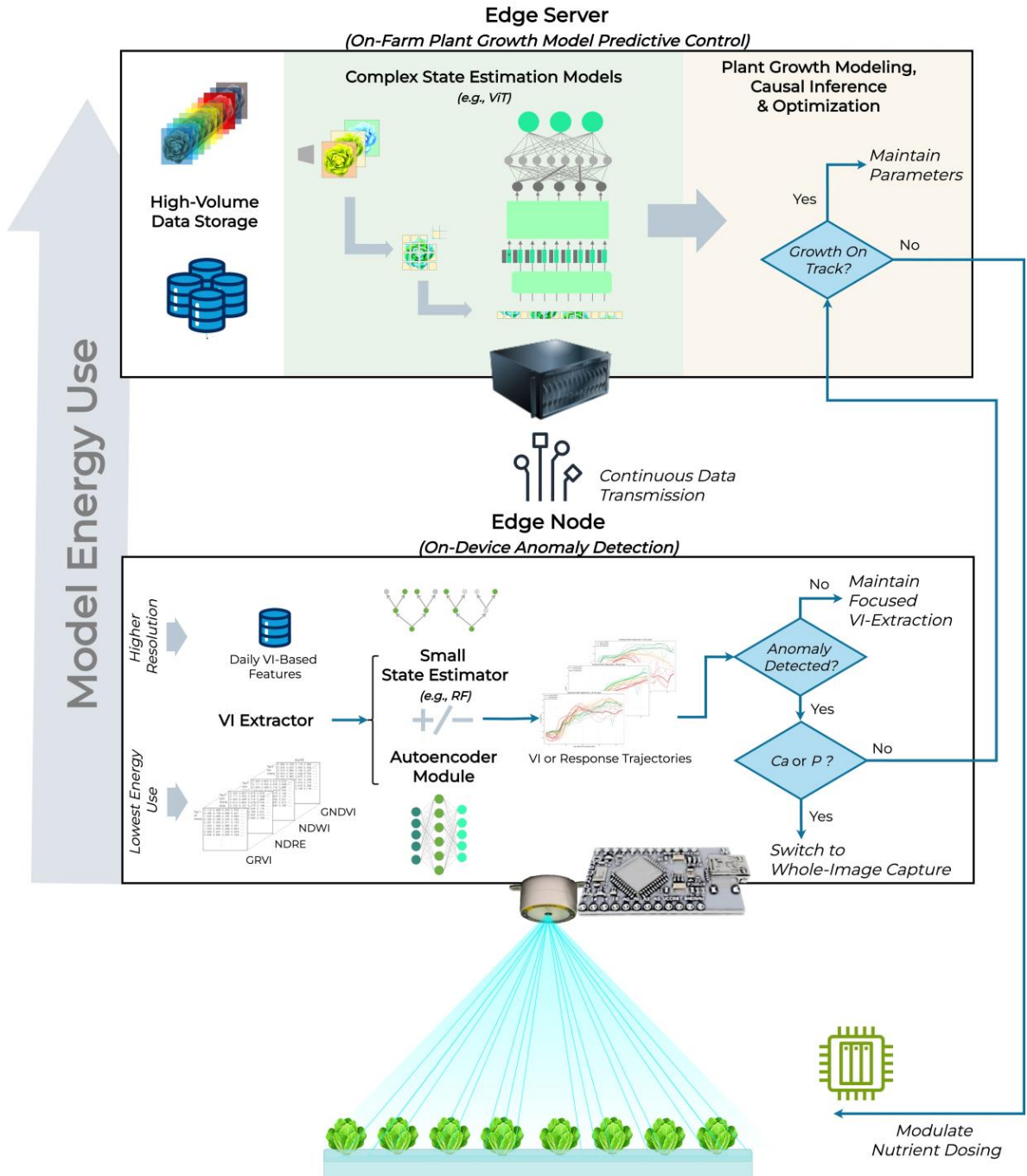

*Figure 5: A proposed end-to-end implementation of the nutrient management automation architecture, including lower-energy modules such as VI-AE and RF-AE, which could be deployed on edge devices, and complex state-estimation approaches such as ViT and future plant growth models, which would be run on edge servers for on-farm MPC.*

In larger farming operations with strict discharge limits, a higher accuracy P detection performance of the ViT may improve operators' compliance at scale. In such operations, energy differences may be justified by increased P-estimation accuracy. Similarly, in a growing environment particularly susceptible to calcium issues and tipburn, such as those using artificial lighting,[75,76] the benefits of the ViT may outweigh the energy costs. By contrast, smaller farming businesses may opt for event-based detection and use deeper analysis or investigation only when triggered. Decoupling the state estimation model from the initial anomaly detection will provide these operators with the ability to monitor deviations to plant status before they become costly mistakes.

Future efforts should optimize RF and AE modules for canopy-level VIs extracted from unsegmented images to adapt the approach for full end-to-end edge device capability. Next, different change point detection architectures should be evaluated for performance and energy use and then optimized based on these results, while the plant state estimation models should be optimized for performance across nutrients with weaker results such as Mg and Ca and with respect to energy use to further the efficiency and sustainability gains. Future work is also needed to extend the system boundary to include CPU and RAM energy use and to evaluate the embodied emissions of computing and controller systems through full life cycle and techno-economic analyses that include the cost of deploying distributed models in a farm environment. Finally, root cause analysis is needed to bridge these early warning systems with causal inference-based plant growth models to close the loop on automated decision support with course correction once deviations are detected.

## 4 CONCLUSIONS

The nutrient monitoring pipeline proposed in this work demonstrates the potential to utilize MSI-measured indicators and feature engineering to dynamically assess nutrient use throughout lettuce growth. We provide a novel contextualization of model energy costs against embodied nutrient waste energy, establishing an adaptable, hierarchical early warning approach to nutrient management. By evaluating different modules for their potential accuracy and efficiency, we provide evidence that ML-enabled nutrient estimation can achieve significant sustainable nutrient management gains. Our results reveal that DL-enabled modeling can substantially reduce negative impacts of fertilizer mismanagement in agriculture, with the most complex model evaluated requiring 40-90x less GPU energy than the embodied energy of wasted N from over application. We demonstrate that use of temporal and feature optimization, combined with nutrient-specific feature engineering for the RF and AE achieves biologically grounded anomaly detection and nutrient estimation without sacrificing accuracy (with an average $R^2$ of 0.81 for FW and N, and net anomaly detection surpassing 80% for severe nutrient deprivation). Further, we show the flexibility of the phenotyping model-agnostic AE module, first as a lightweight early warning system that triggers further analysis only when necessary, then a more integrated system-level tool when combined with a state estimation module for higher-resolution trajectory analysis. This modular approach has the potential to improve operational efficiency and adaptability in a range of environments and operating conditions.

This research has broad implications for deployment and discovery. First, it supports phenology research to uncover dynamic uptake patterns and dose response in agriculture with potential to facilitate the optimization of growth parameters that are currently not possible with destructive sampling workflows. Next, the flexible VI-based AE module can be adapted for deployment in combination with increasingly high-resolution phenotyping models as part of alternative automation pipelines to save operators time and labor costs associated with high-frequency manual spot checks. Given the broad indications of VI-based features, this has potential applicability for pest management, water use optimization, and more. This integrated analysis of model accuracy and energy efficiency provides a framework for resource-aware nutrient management and sustainability in agriculture.

**Abbreviations**

***Controlled Environment Agriculture (CEA):*** Agricultural systems within enclosed structures such as greenhouses and vertical farms where environmental conditions are monitored and controlled for optimal plant growth.

***Precision Agriculture (PA):*** Farming management approach using data and technology to optimize agricultural inputs and operations for improved efficiency and sustainability.

***Vegetation Index (VI)***: Numerical indices calculated over the foliar surface by relating different spectral band reflectivity intensities.

***Random Forest (RF)***: Machine learning technique using layers of decision trees for classification or regression tasks.

***Vision Transformer (ViT):*** Deep learning architecture that processes images as sequences of patches using self-attention mechanisms.

***Autoencoder (AE):*** Deep learning architecture specialized for unsupervised learning of latent representations and reconstruction.

***RGB (Red, Green, Blue):*** Color imaging system capturing visible light in three channels corresponding to red, green, and blue wavelengths.

***Multispectral Imaging (MSI)***: Imaging technique capturing several discrete wavelengths of visual and near infrared reflectance data for each spatial pixel.

***Deep Learning (DL):*** Subset of machine learning using neural networks with multiple layers to learn hierarchical representations.

**Acknowledgements**

This research was funded in part by U.S. Department of Agriculture (Award Nos. 2018-68011-28371, 2021-67021-34499, 2021-67021-38585, and 2024-67021-41534); National Science Foundation (Award Nos. 2112533, 2345543, and 2419122). Special thanks to the student volunteers that made this research possible: Liam Nunn, Swathi Mugundu Pradeep, Bronwyn Armitage, and Jackson Martin. Authors would also like to acknowledge the use of Claude Sonnet 4 (claude.ai), ChatGPT4 (chatgpt.com), and Visual Studio Copilot (code.visualstudio.com) for coding assistance and debugging. Code snippets generated were thoroughly reviewed, customized, and validated by the authors during this process, and authors take full responsibility for the content of this study.

**CRediT Author contribution statement**

**Abigail Cohen:** Writing – review and editing, Writing – original draft, Visualization, Validation, Methodology, Formal Analysis, Data curation, Conceptualization. **Yuming Sun:** Writing – review and editing, Writing – original draft, Visualization, Methodology, Formal analysis. **Zhihao Qin:** Software, Investigation, Formal analysis. **Harsh Muriki:** Writing – review and editing, Writing – original draft, Visualization, Validation, Methodology, Formal Analysis. **Yeonju Lee:** Writing – review and editing, Methodology, Conceptualization. **Zihao Xiao:** Writing – review and editing, Formal analysis. **Matt Housley:** Investigation. **Andrew Sharkey:** Investigation. **Rhuanito Soranz Ferrarezi:** Writing – review and editing, Resources. **Jing Li:** Writing – review and editing, Conceptualization, Supervision. **Lu Gan:** Writing – review and editing, Supervision. **Yongsheng Chen:** Writing – review and editing, Supervision, Funding acquisition, Data curation, Conceptualization.

**Declaration of Competing Interests**

The authors have no competing interests, financial or otherwise, to declare.

**Data Availability Statement**

Our plant tissue data and vegetation index calculations will be available at https://agdatacommons.nal.usda.gov/, status pending. Images will be made available upon request.

**Funding**

This study was partially supported by U.S. Department of Agriculture (Award Nos. 2018-68011-28371, 2021-67021-34499, 2021-67021-38585, and 2024-67021-41534); National Science Foundation (Award Nos. 2112533, 2345543, and 2419122).